\documentclass[10pt]{article} 
\usepackage{graphicx}
\usepackage[preprint]{tmlr}

\usepackage{amsmath,amsfonts,bm}

\def\eqref#1{equation~\ref{#1}}

\def\1{\bm{1}}

\DeclareMathAlphabet{\mathsfit}{\encodingdefault}{\sfdefault}{m}{sl}
\SetMathAlphabet{\mathsfit}{bold}{\encodingdefault}{\sfdefault}{bx}{n}

\usepackage{hyperref}
\usepackage{url}

\title{Training Large Language Models to Predict Clinical Events}

\author{\name Benjamin Turtel\thanks{\texttt{Contact: ben@lightningrod.ai}}, \ Paul Wilczewski, \ Kris Skotheim \\
      \addr Lightning Rod Labs}

\def\month{MM}  
\def\year{YYYY} 
\def\openreview{\url{https://openreview.net/forum?id=XXXX}} 

\begin{document}

\maketitle

\begin{abstract}
Longitudinal clinical notes contain rich evidence of how patients evolve over time, but converting this signal into training supervision for clinical prediction remains challenging. We extend Foresight Learning to clinical prediction by converting time-ordered MIMIC-III notes into examples consisting of past patient context, a natural-language question about a possible future event, and a label resolved from later documentation. This process yields 6,900 prediction examples from 702 admissions across medications, procedures, organ support, microbiology, and mortality. A small LoRA adapter trained on these examples improves over the prompted base model, reducing ECE from 0.1269 to 0.0398 and Brier score from 0.199 to 0.145, and slightly outperforming GPT-5 point estimates on held-out questions. The approach enables reusable clinical prediction supervision from longitudinal notes without hand-engineered structured features or endpoint-specific classifiers.
\end{abstract}

\section{Introduction}

Clinical decision-making often depends on anticipating how a patient’s condition will evolve from incomplete and continuously updated information. Electronic health records contain longitudinal evidence of patient status, treatment response, and clinician assessment, but much of this signal is captured in free-text notes rather than fixed structured variables or static labels.

In this work, we extend our prior work on Foresight Learning~\cite{foresightlearning2026} to clinical event predictions. The key observation is that earlier notes define what was known at a prediction time, while later documentation records how the patient’s condition evolved and which outcomes occurred. Our main contribution is an end-to-end framework for turning raw, time-ordered EHR notes into clinical prediction models. We operationalize this idea by constructing chronological trajectories from MIMIC-III notes, sampling prediction times, generating natural-language questions about possible future events, and resolving those events using subsequent clinical evidence. 

This process produces event-specific prediction examples consisting of a partial patient history, a question about a possible future outcome, and a label resolved from the later clinical record. Because the questions are expressed in natural language, the same trajectory can support heterogeneous predictions across medications, procedures, organ support, microbiology results, and mortality, without requiring endpoint-specific classifiers. 

We then use this longitudinal supervision to adapt an open-weight 120B-parameter language model with a small LoRA adapter, producing a specialized probabilistic clinical prediction model without full-parameter fine-tuning. In a retrospective demonstration on MIMIC-III, the adapted model substantially improves over the prompted base model and slightly exceeds GPT-5 point estimates under the same benchmark setup.

\section{Related Work}

\subsection{Longitudinal EHR Prediction}

A large body of work uses electronic health records to predict mortality, readmission, diagnosis progression, and future clinical events. Earlier approaches relied on structured variables and hand-engineered risk scores, while more recent work models patient histories directly with sequence models and transformers. BEHRT~\cite{behrt2020} applies transformer architectures to longitudinal EHR records for disease prediction, and Med-BERT~\cite{medbert2021} learns contextualized representations from structured patient trajectories. More recent patient-timeline systems extend this direction with richer temporal modeling: Foresight 2~\cite{foresight22024} constructs patient timelines from MIMIC-III notes using biomedical concept extraction and fine-tunes open models for diagnosis prediction, medication recommendation, and risk forecasting, while GRAIL~\cite{grail2026} studies trajectory prediction from structured patient histories with hierarchical embeddings and LLM reranking on MIMIC-IV.

These approaches demonstrate the value of modeling patients over time, but generally rely on structured codes, extracted biomedical concepts, or next-event prediction rather than explicit natural-language questions over raw clinical narratives. This matters because much of the clinically relevant signal lives in unstructured notes - nuanced assessments, evolving reasoning, and findings that resist easy quantification.

\subsection{Language Models for Clinical Notes}

Another line of work applies language models directly to clinical text. ClinicalBERT~\cite{clinicalbert2019} showed that domain-adapted transformers trained on clinical notes can improve downstream hospital prediction tasks such as readmission and phenotype classification. More broadly, physician and nursing notes contain predictive information that is not fully captured by structured variables.

Most note-based systems treat documentation as static input for classification, extraction, or summarization. In contrast, our setup treats clinical notes as an evolving record: the model sees what was documented up to a prediction time and predicts outcomes resolved later.

\subsection{Foresight Learning}

This work builds on Foresight Learning~\cite{foresightlearning2026}, a framework for training models to make probabilistic predictions using only information available at prediction time, with supervision derived from later realized outcomes. Prior applications of this framework have demonstrated its utility in forecasting SEC risks~\cite{secrisks2026} and supply chain disruptions~\cite{supplychain2026}, showing that temporally grounded supervision can support prediction in complex, real-world domains where future outcomes must be inferred from evolving textual and structured evidence.

We apply this framework to clinical narratives by constructing patient trajectories from MIMIC-III notes, generating prediction question-answer pairs from those trajectories, and fine-tuning a model to produce calibrated clinical predictions. Taken together, prior work shows that structured EHR histories, clinical notes, and task-specific fine-tuning each improve healthcare prediction, while recent Foresight Learning applications suggest that models can learn predictive behavior from outcome-resolved, temporally grounded examples across domains. Our contribution is to combine these ideas in a clinical setting: transforming unstructured clinical narratives into temporally grounded question-answer pairs and training a model to predict future outcomes using only the information available at each point in the patient trajectory.

\section{Data and Problem Setup}

\begin{figure}[ht]
    \centering
    \includegraphics[width=\linewidth]{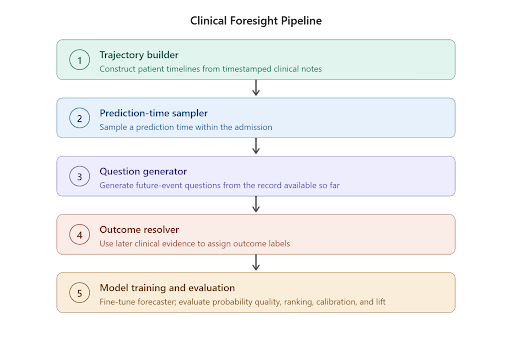}
    \caption{Clinical prediction pipeline.}
    \label{fig:pipeline}
\end{figure}

Figure~\ref{fig:pipeline} summarizes the data construction and evaluation pipeline. Starting from timestamped clinical notes, we construct patient trajectories, sample prediction times, generate prediction questions from the record available at those times, resolve outcomes using later clinical evidence, and use the resulting examples for model training and evaluation.

\subsection{Data Source and Trajectory Construction}

Our analysis uses MIMIC-III v1.4, a de-identified critical care dataset made available through PhysioNet for credentialed research use \cite{mimic2016}. Access was obtained under the required PhysioNet credentialing and MIMIC data use terms. All model-based processing of MIMIC-III notes, including question generation, label resolution, evaluation, and training, was performed using providers and environments confirmed to comply with the applicable data use requirements.

MIMIC-III contains hospital admissions, ICU stays, demographics, diagnoses, procedures, medications, laboratory measurements, charted events, and longitudinal clinical notes. For each hospital admission, we construct a chronological patient trajectory by ordering all available free-text notes by timestamp. These notes include nursing documentation, physician progress notes, consult notes, radiology interpretations, discharge summaries, and other narrative records. The resulting trajectory represents the evolving clinical record of a patient over the course of an admission: earlier notes capture what was known at the time, while later notes document subsequent interventions, outcomes, and clinical developments. To preserve temporal realism, trajectories are constructed strictly in timestamp order.

\subsection{Question Construction}

We convert each retrospective patient trajectory into prediction examples by randomly selecting a single split time strictly before the recorded discharge time. Notes available up to the split define the prediction context, while discharge information and subsequent clinical evidence are withheld from the model input and used only for outcome resolution.

For each trajectory, we use Gemini 2.5 Flash to generate multiple clinically meaningful prediction questions conditioned only on documentation available before the split time. The model is instructed to ask about plausible future events during the remainder of the same admission, such as medication initiation, procedures, organ support, laboratory or microbiology findings, and mortality. The question-generation model does not receive post-split notes or discharge documentation. Generated questions are then resolved separately using post-split documentation, and questions that cannot be assigned a supported binary label are excluded.

The resulting questions target observable future events documented in the remainder of the admission, including medication or therapy initiation, procedures and organ support, laboratory or microbiology results, and mortality. Representative examples include:

\begin{itemize}
    \item Will the patient be started on intravenous vasopressors during this admission?
    \item Will the patient receive a blood transfusion of packed red blood cells during this admission?
    \item Will the patient receive renal replacement therapy (dialysis) during this admission?
    \item Will the patient be declared dead during this hospital admission?
    \item Will the patient require endotracheal intubation for mechanical ventilation during this admission?
    \item Will the patient's sputum culture return positive for a pathogenic bacterial or fungal organism during this admission?
\end{itemize}

\subsection{Label Resolution and Prediction Task}

Each generated question is resolved using only documentation after the split time from the same admission, including discharge documentation when available. Gemini 2.5 Flash assigns a binary label based on whether the future record contains sufficient evidence that the queried event occurred after the prediction time and before discharge.

Because question generation and label resolution occur on opposite sides of the trajectory split, each example is grounded in a realistic setup: the question is based only on information available at prediction time, and the answer is determined only from later clinical evidence. This avoids look-ahead bias and mirrors the setting we want the model to learn.

Formally, for patient $i$, split time $t$, and future clinical event $e$, we define
\begin{equation*}
    y_{i,t,e} = \mathbf{1}\{\text{event } e \text{ occurs after } t \text{ and before discharge}\}.
\end{equation*}
The model is trained to estimate
\begin{equation*}
    P\left(y_{i,t,e}=1 \mid \mathrm{trajectory}_{i,\leq t}\right).
\end{equation*}

That is, the probability that a specified clinical event occurs later in the admission, given only the clinical notes available up to the split. The model outputs a numerical probability and may optionally provide natural-language reasoning grounded in the observed trajectory.

To make the setup concrete, Table~\ref{tab:synthetic-example} shows a simplified synthetic example based on one of the event types used in the dataset. The example is illustrative only and does not reproduce any actual patient text from MIMIC-III, in compliance with the data use agreement.

\begin{table}[t]
    \centering
    \caption{Synthetic temporally grounded clinical prediction example.}
    \label{tab:synthetic-example}
    \begin{tabular}{p{0.25\linewidth}p{0.65\linewidth}}
        \hline
        \textbf{Component} & \textbf{Synthetic example} \\
        \hline
        Prediction time & $T=0$, 48 hours after admission \\
        Observed record at prediction time & The patient is admitted with severe pneumonia and worsening oxygen needs. Notes describe increasing work of breathing, persistent hypoxemia despite supplemental oxygen, and concern that respiratory status may deteriorate. The patient has not yet been intubated at the prediction time. \\
        Question & Will the patient require endotracheal intubation for mechanical ventilation during this admission? \\
        Resolution time & $T=\text{outcome}$, end of admission \\
        Later clinical evidence & Subsequent notes document worsening respiratory failure overnight. The patient is intubated and placed on mechanical ventilation during the same admission. \\
        \hline
    \end{tabular}
\end{table}

\subsection{Dataset Statistics and Splits}

The final dataset consists of prediction questions generated from a random sample of hospital admissions with sufficient longitudinal documentation. This dataset represents a sampled subset of the available note-derived supervision rather than an exhaustive enumeration: the same pipeline can generate additional examples by sampling more admissions, selecting additional split times within each trajectory, or generating more questions per split. We include admissions with at least nine timestamped notes and a recorded discharge time, ensuring that each trajectory contains sufficient longitudinal context and a well-defined endpoint for label resolution. For each included admission, we construct one chronological trajectory, select one split time before discharge, generate multiple prediction questions from the observed portion of the trajectory, and assign labels using the remaining future record.

We partition the dataset at the admission level so that no admission appears in more than one split. The held-out test set contains 500 questions, and all remaining questions are used for training. The resulting dataset contains 702 admission-level trajectories and 6,900 questions, with an average of 9.8 questions per trajectory and a positive label rate of 25

\begin{table}[t]
    \centering
    \caption{Dataset summary.}
    \label{tab:dataset-summary}
    \begin{tabular}{lr}
        \hline
        \textbf{Statistic} & \textbf{Value} \\
        \hline
        Admissions / trajectories & 702 \\
        Mean questions per trajectory & 9.8 \\
        Positive label rate & 25\% \\
        Train questions & 6,400 \\
        Test questions & 500 \\
        \hline
    \end{tabular}
\end{table}

\section{Model}

\subsection{Learning Framework and Architecture}

We formulate clinical event prediction as conditional probabilistic predictions. Given the clinical record available at a prediction time and a natural-language question about a future clinical event, the model estimates the probability that the event occurs later in the admission.

Our base model is gpt-oss-120b, a 120B-parameter decoder-only language model. We adapt it using Low-Rank Adaptation (LoRA) with rank r=32, keeping the base weights frozen and training only task-specific adapter parameters. Training is performed through the Lightning Rod SDK, which supports multiple backend training engines. We use Tinker as our backend for this work. This enables efficient specialization to the clinical prediction task without full-parameter fine-tuning.

Each input contains a task instruction, the chronological patient record available at the prediction time, and the prediction question. Inputs are truncated to a maximum context length of 16,000 tokens, preserving the most recent clinical documentation when the available record exceeds this limit. The model outputs a numerical probability between 0 and 1, interpreted as the estimated likelihood that the queried event occurs after the prediction time and before discharge.

Because each patient record can be paired with multiple event-specific questions, the same model can predict heterogeneous outcomes, including medication initiation, procedures, organ support, microbiology results, and mortality. Thus, the model is not trained as a separate classifier for each endpoint, but as a general event-conditioned prediction model. This general-purpose design also confers robustness to variation in input context: the model adapts to records that differ in length, documentation style, and available information, without requiring a fixed or standardized input format. 

\subsection{Training Objective and Optimization}

We train the model under the Foresight Learning framework, using realized clinical outcomes to reward predictions made from the information available at prediction time. Following our prior Foresight Learning~\cite{foresightlearning2026} work, the model is optimized to produce both a probability estimate and a reasoning trace supporting its prediction.

The primary reward is the log score, a proper scoring rule for probabilistic forecasts. For predicted probability $p$ and realized binary outcome $y \in \{0,1\}$, the reward is defined as

\begin{equation*}
    r = y \log p + (1-y)\log(1-p).
\end{equation*}

This objective rewards predictions that assign high probability to the realized outcome and penalizes overconfident errors. Maximizing expected log score is equivalent to maximizing the likelihood of observed outcomes under the model’s predictive distribution.

We optimize the LoRA adapters using GRPO with group size of 4 and a batch size of 32. For each example, the model samples four full reasoning traces and probability estimates, each of which is scored against the realized binary outcome using the log-score reward. Only the LoRA adapter parameters are updated; the base model weights remain frozen.

This training procedure encourages the model not only to output calibrated probabilities, but also to produce reasoning traces grounded in the available clinical record. Reported results use the final checkpoint selected based on validation performance.

\section{Results}

We evaluate model performance on held-out questions constructed from patient trajectories. Test examples are separated from training data at both the admission ID and patient ID levels to prevent leakage across admissions or patients. At prediction time, all models receive the same clinical record and question.

\subsection{Baselines and Metrics}

We compare the fine-tuned prediction model against three reference points: a constant-probability baseline that predicts the training-set positive label rate for every example, the prompted gpt-oss-120b base model without task-specific fine-tuning, and GPT-5 as a general-purpose external benchmark. Together, these comparisons distinguish the contribution of event prevalence, the prior predictive ability of the base model, and the effect of task-specific adaptation.

Performance is assessed using reward, Brier score, expected calibration error, AUROC, and top-10\% lift. Reward is the log-score objective used during training, with higher values indicating better predictions. Brier score and expected calibration error measure probability quality and calibration, while AUROC measures ranking performance. Top-10\% lift measures enrichment among the model’s highest-risk predictions: the event rate in the top 10\% of predicted probabilities relative to the overall event rate. A value above 1 indicates that positive outcomes are concentrated among the model’s highest-risk predictions.

\subsection{Aggregate Performance}

Table~\ref{tab:aggregate-performance} summarizes overall performance on the held-out test set. The main result is that a lightweight adapter trained with Foresight Learning turns a prompted open model into a substantially stronger clinical prediction model. The trained gpt-oss-120b adapter improves over the prompted base model across every reported metric and performs competitively with GPT-5, slightly exceeding its point estimates under the same retrospective benchmark setup.

The gains appear in both probability quality and ranking performance. Relative to the prompted base model, reward improves from -0.5856 to -0.4586, Brier score decreases from 0.1994 to 0.1453, ECE decreases from 0.1269 to 0.0398, AUROC rises from 0.6992 to 0.7993, and top-10\% lift rises from 2.34 to 3.07. These improvements indicate that task-specific adaptation improves both calibration and risk ranking.

Top-10\% lift is especially relevant for applications where only a limited number of high-risk cases may be reviewed. The trained model’s top-10\% lift of 3.07 means that the highest-risk decile contains positive outcomes at roughly three times the overall event rate.

\begin{figure}[htbp]
    \centering
    \includegraphics[width=\linewidth]{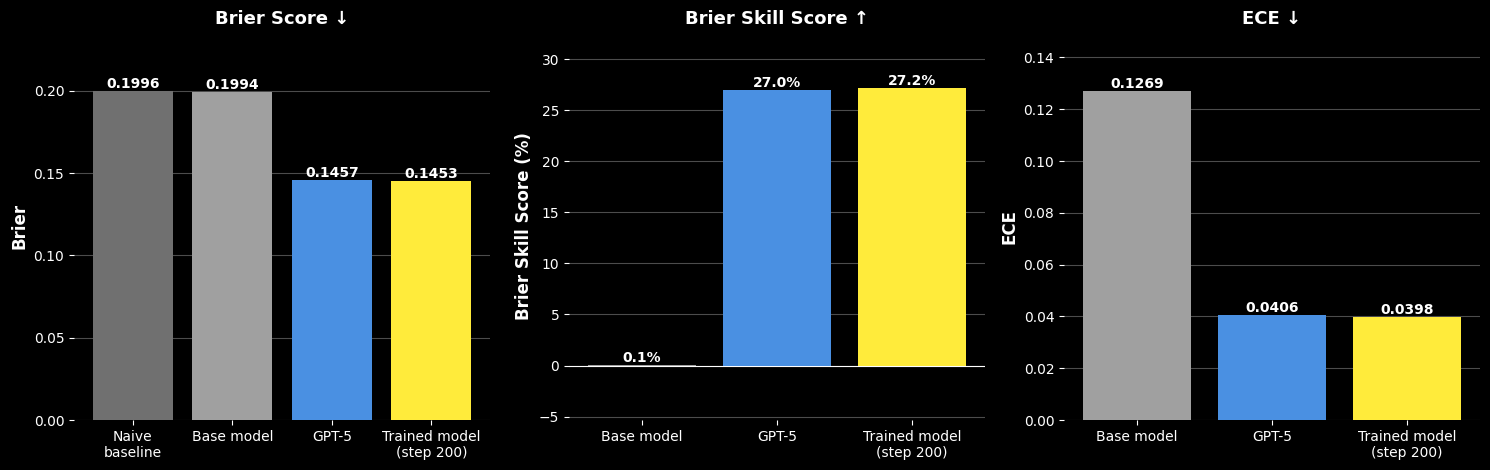}
    \caption{Test set performance by model.}
    \label{fig:test-performance}
\end{figure}

\begin{table}[htbp]
    \centering
    \caption{Aggregate probabilistic performance on held-out test set.}
    \label{tab:aggregate-performance}
    \begin{tabular}{lrrrrr}
        \hline
        \textbf{Model} & \textbf{Reward} & \textbf{Brier} & \textbf{ECE} & \textbf{AUROC} & \textbf{Top-10\% lift} \\
        \hline
        Base model & -0.5856 & 0.1994 & 0.1269 & 0.6992 & 2.3358 \\
        GPT-5 & -0.4636 & 0.1457 & 0.0422 & 0.7954 & 2.9927 \\
        Trained Model (step 200) & -0.4586 & 0.1453 & 0.0398 & 0.7993 & 3.0657 \\
        Constant baseline (24.8\%) & -0.5890 & 0.1996 & & & \\
        \hline
    \end{tabular}
\end{table}

A reliability diagram further illustrates calibration differences between the prompted and fine-tuned models. The fine-tuned model's predicted probabilities more closely track empirical event frequencies across probability bins, while the prompted base model is less well calibrated.

\begin{figure}[htbp]
    \centering
    \includegraphics[width=0.5\linewidth]{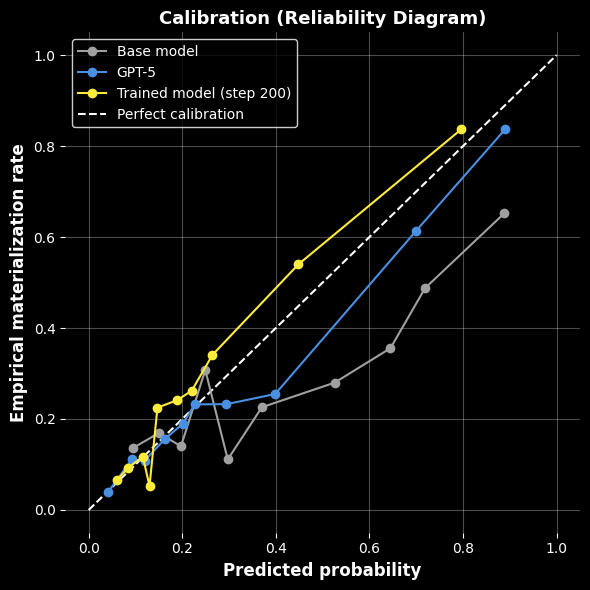}
    \caption{Reliability diagram comparing prompted and fine-tuned models.}
    \label{fig:reliability}
\end{figure}

\subsection{Reasoning Quality Comparison}

To better understand behavioral differences, we reviewed 50 matched prediction examples from the prompted base model and the fine-tuned model. Because the reviewed examples contain sensitive clinical text, we report only aggregate qualitative observations rather than patient-level examples.

Using Gemini 2.5 Flash as an impartial judge, we employed a blind evaluation approach to assess qualitative differences in model reasoning. For each matched pair, the evaluator model was presented with outputs from both systems in randomized order, without any label indicating which system produced each response, and asked to identify the superior response across four predefined dimensions: clinical reasoning, medical knowledge, grounding, and clinical utility. In this review, we find the fine-tuned model more often incorporated temporally relevant clinical evidence, connected patient-specific findings to the predicted outcome, and considered alternative future scenarios when expressing uncertainty.

Compared with the base model, the fine-tuned model's reasoning was generally more detailed and more explicitly tied to the patient's evolving clinical course. As shown in Table 4, the trained model outperformed the base model across all evaluated dimensions, with the largest margins in medical knowledge (92.0\%) and grounding (78.0\%), and an overall win-rate of 84.0\%. These results are consistent with the quantitative improvements in calibration and prediction performance, and suggest that outcome-based training improves the model's ability to produce clinically grounded reasoning for its predictions.

\begin{table}[h]
\centering
\caption{LLM-judge win-rates by evaluation dimension.}
\label{tab:llm_judge}
\begin{tabular}{lc}
\hline
\textbf{Dimension} & \textbf{Trained model win-rate} \\
\hline
Clinical reasoning & 78.0\% \\
Medical knowledge  & 92.0\% \\
Grounding          & 78.0\% \\
Clinical utility   & 82.0\% \\
\hline
Overall            & 84.0\% \\
\end{tabular}
\end{table}

\section{Discussion}

\subsection{Interpretation of Results}

Our results show that outcome-based training on temporally grounded prediction questions can meaningfully improve clinical prediction performance. The fine-tuned model substantially improves over the prompted gpt-oss-120b base model and slightly outperforms GPT-5 across metrics. This supports the central claim of this work: models can learn specialized predictive behavior from retrospective patient trajectories.

This result is notable given the messiness of real EHR data. Clinical notes often include autofilled text, templated language, repeated documentation, and other artifacts that may carry limited signal for a given prediction. Because the model is trained on outcome-resolved trajectories rather than manually selected features, this objective allows it to learn from the full record without specifying in advance which note types or text patterns should matter.

\subsection{Contributions of the Data Construction Framework}
A central contribution of this work is a reusable workflow for converting completed clinical trajectories into LLM-ready training and evaluation examples. Earlier documentation defines the prediction context, while later evidence resolves the outcome. This allows the same evolving patient record to support heterogeneous natural-language prediction while preserving realistic information constraints.

This approach scales across many admissions, supports multiple outcome types through a single question-conditioned interface, and grounds supervision in observed patient trajectories. The same framework could be adapted to other EHR datasets, specialty-specific registries, outpatient records, or multimodal patient timelines that combine notes, labs, medications, procedures, and imaging reports.

More broadly, hospitals, disease registries, and specialty cohorts already contain longitudinal records of patient care. This method converts those records into training signal for many future-event predictors without requiring manual endpoint labeling or separate hand-built classifiers for each outcome. A registry of liver cancer patients, a hospital ICU population, or a demographic-defined cohort could each support a model attuned to the outcomes and signals most relevant to that group.

\subsection{Limitations and Future Work}

These results should be interpreted in the context of several constraints. First, MIMIC-III is a single-center retrospective dataset and may not reflect other institutions, patient populations, or current practice patterns. Second, clinical notes are noisy, incomplete, and shaped by documentation behavior, which may limit signal quality and introduce bias. Third, both prediction-question generation and outcome resolution rely on automated model-based processing, introducing possible errors in question relevance, label assignment, or event timing. Fourth, the model presented here is trained solely on MIMIC-III notes and is restricted by the MIMIC data use terms to scientific research use; it should be understood as a demonstration of the approach rather than a deployable clinical product. Fifth, neither GPT-5 nor gpt-oss-120b disclose their pretraining data, and either may have been trained on MIMIC-III or derived works, limiting the ability to assess true out-of-distribution generalization.

Future work should apply the framework to additional patient trajectory datasets, integrate structured EHR data with narrative notes, and audit generated questions and resolved labels more rigorously. Another important direction is the development of shared benchmarks for temporally grounded clinical predictions, enabling researchers to compare methods across datasets, outcome types, and prediction horizons.

\section{Conclusion}

We presented an end-to-end application of Foresight Learning for training clinical prediction models from longitudinal EHR notes. By using earlier documentation as prediction context and later documentation as outcome evidence, we derived temporally grounded supervision from completed patient trajectories and used it to adapt gpt-oss-120B with a lightweight LoRA adapter. In a retrospective demonstration on MIMIC-III, the resulting model substantially improved over the prompted base model and slightly outperformed GPT-5 under the same benchmark setup.

More broadly, these results suggest that longitudinal clinical records can support future-event prediction directly from evolving patient context. The same routine documentation that clinicians already produce can be used to train models that anticipate what patients may need next. In practical terms, health systems can turn the clinical data they already have into AI models that make useful predictions.

\newpage

\bibliography{main}
\bibliographystyle{tmlr}

\end{document}